\documentclass[sigconf]{acmart}
\usepackage{multirow}
\usepackage[ruled,vlined, linesnumbered]{algorithm2e}

\AtBeginDocument{%
  \providecommand\BibTeX{{%
    \normalfont B\kern-0.5em{\scshape i\kern-0.25em b}\kern-0.8em\TeX}}}

\setcopyright{acmcopyright}
\copyrightyear{2021}
\acmYear{2021}
\acmDOI{10.1145/nnnnnnn.nnnnnnn}

\acmConference[GECCO '21]{the Genetic and Evolutionary Computation Conference 2021}{July 10--14, 2021}{Lille, France}
\acmPrice{15.00}
\acmISBN{978-x-xxxx-xxxx-x/YY/MM}

\begin{document}

\title[Which Hyperparameters to Optimise?]{Which Hyperparameters to Optimise? An Investigation of Evolutionary Hyperparameter Optimisation in Graph Neural Network For Molecular Property Prediction}


\author{Yingfang Yuan}
\affiliation{%
  \institution{Heriot-Watt University}
  \streetaddress{First Gait}
  \city{Edinburgh}
%   \state{Scotland} 
  \country{United Kingdom}
  \postcode{EH14 4AS}}
\email{yyy2@hw.ac.uk}

\author{Wenjun Wang}
\affiliation{%
  \institution{Heriot-Watt University}
%   \streetaddress{First Gait}
  \city{Edinburgh}
%   \state{Scotland} 
  \country{United Kingdom}
  \postcode{EH14 4AS}}
\email{wenjun.wang@hw.ac.uk}

\author{Wei Pang}
\authornote{Corresponding author}
\affiliation{%
  \institution{Heriot-Watt University}
%   \streetaddress{First Gait}
  \city{Edinburgh}
%   \state{Scotland}
  \country{United Kingdom}
  \postcode{EH14 4AS}}
\email{w.pang@hw.ac.uk}

\renewcommand{\shortauthors}{}

\begin{abstract}
Recently, the study of graph neural network (GNN) has attracted much attention and achieved promising performance in molecular property prediction. Most GNNs for molecular property prediction are proposed based on the idea of learning the representations for the nodes by aggregating the information of their neighbor nodes (e.g. atoms). Then, the representations can be passed to subsequent layers to deal with individual downstream tasks. Therefore, the architectures of GNNs can be considered as being composed of two core parts: graph-related layers and task-specific layers. Facing real-world molecular problems, the hyperparameter optimization for those layers are vital. Hyperparameter optimization (HPO) becomes expensive in this situation because evaluating candidate solutions requires massive computational resources to train and validate models. Furthermore, a larger search space often makes the HPO problems more challenging. In this research, we focus on the impact of selecting two types of GNN hyperparameters, those belonging to graph-related layers and those of task-specific layers, on the performance of GNN for molecular property prediction. In our experiments. we employed a state-of-the-art evolutionary algorithm (i.e., CMA-ES) for HPO. The results reveal that optimizing the two types of hyperparameters separately can gain the improvements on GNNs' performance, but optimising both types of hyperparameters simultaneously will lead to predominant improvements. Meanwhile, our study also further confirms the importance of HPO for GNNs in molecular property prediction problems.
\end{abstract}





\maketitle

% \section{Introduction}
\section{Introduction}
Graph neural networks (GNNs) have been applied to solve a wide range of problems, such as social recommendation \cite{fan2019graph}, citation trend prediction \cite{cummings2020structured}, and molecular property prediction \cite{yuan2021systematic, duvenaud2015convolutional, gilmer2017neural, jiang2021could}. One advantage of GNNs is that they can be directly operated on graphs in an end-to-end manner for real-world problems, and task-specific representations can be learned automatically between latent layers. In contrast, traditional machine learning methods require handcrafted features from graph as input. For example, the work presented in~\cite{jiang2021could} mentioned that support vector machine, random forest, and XGBoost took handcrafted descriptors and/or fingerprints from molecular structures as inputs to predict molecular properties. When GNNs are applied to the same problems, given a molecular graph, the vector representations for all atoms are learned at first by aggregating and updating the information of neighbor atoms. Thereafter, the readout operator \cite{wu2020comprehensive} (i.e., mean, sum) is employed to collapse all atom representations into a molecular (graph-level) representation, which can be passed to subsequent task-specific layers to make predictions. In our experiments, we employed graph convolution \cite{duvenaud2015convolutional} because it has been proposed considering molecular domain knowledge.

The study presented in \cite{yuan2021systematic, yuan2021novel, yuan2021genetic} indicated that hyperparameter optimization (HPO) by evolutionary computation can improve the GNN's performance for predicting molecular properties. However, HPO for GNNs is often an expensive task. Evaluating hyperparameter settings on GNNs needs to train models which cost a lot of computational resources. Some methods have been proposed to alleviate this issue such as using surrogate models \cite{bergstra2011algorithms}, successive halving \cite{karnin2013almost, jamieson2016non}. From another angle to consider this issue, the predefined hyperparameter search space may directly affect the HPO results. For example, it is very challenging to search for optimal solutions given a very large hyperparameter space and a relatively limited computational budget. The HPO search space is defined by selected hyperparameters (e.g., batch size, learning rate), the ranges of these hyperparameter values, and the step sizes (i.e, intervals) which determine the resolutions of the hyperparameter values to be searched. From our perspective, the ranges and step sizes depend on individual practical cases to some degree. This research focuses on investigating the impact of selecting different types of hyperparameters to optimize on the GNN's performance for molecular property predictions. In concrete, we investigated the differences of optimizing hyperparameters related to graph layers and those from task-specific layers. Therefore, we grouped the hyperparameters according to the layer-wise neural architectures of GNNs, which can be considered as being the combination of graph layers and task-specific layers. In this way, we expect to discover that optimizing which types of the hyperparameters may bring more expected gains.

Additionally, CMA-ES \cite{hansen2016cma} as a state-of-the-art, derivative-free, and evolutionary black-box optimization method has been employed to optimize hyperparamters. Therefore, we use it as our evolutionary search strategy for HPO. This will allow us to focus on the impact of the two types of hyperparameters with the same search strategy. However, we also point out that more evolutionary HPO approaches could be used in the future to test the impact of hyperparameters, but this is beyond the scope of this research.  . 

The rest of this paper is organized as follows. Section 2 introduces related work. In Section 3, the experiments are reported and the results are analysed. Finally, Section 4 concludes the paper and explores some directions in future work.

\section{Related Work}

\subsection{Graph Neural Networks}
\begin{figure*}
    \centering
    \includegraphics[scale=0.35]{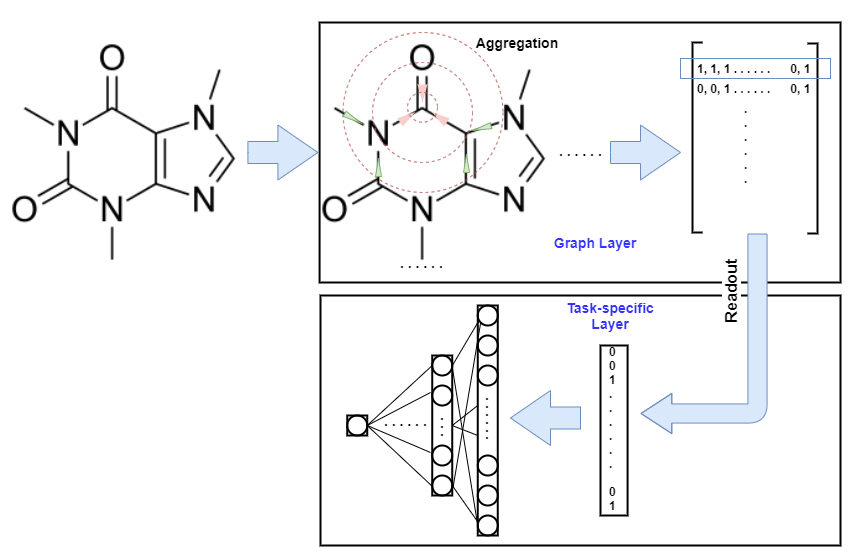}
    \caption{The Neural Architecture of GNN in Molecular Property Prediction \\
\small The input is SMILES \cite{weininger1988smiles} which can be used to describe molecular structure in the form of a line notation}
    \label{fig:gnnmol}
\end{figure*}

Many types of GNNs have been proposed based on the idea of aggregating neighborhood recursively \cite{xu2018powerful} (or message passing \cite{gilmer2017neural}). Each node aggregates feature vectors of its neighbors to compute its new feature representation. To generate the representation of the entire graph, many schemes have been proposed such as graph pooling \cite{ying2018hierarchical}, attention sum \cite{li2015gated}, mean \cite{atwood2016diffusion}, max \cite{gori2005new}. When GNNs are applied in molecular property prediction, the representation of the entire graph is then passed to subsequent layers \cite{jiang2021could}. For example, graph convolution model (GC) \cite{duvenaud2015convolutional} makes use of neural networks to imitate circular fingerprint (a representation of molecular structures by encoding the information of atom neighborhoods.)~\cite{glen2006circular} to generate molecular representation, the so called neural fingerprint, which can be passed to a simple neural network to predict drug efficiency and photovoltaic efficiency.

Therefore, in general, the architectures of GNNs are classified into graph layers and task-specific layers (Fig. \ref{fig:gnnmol}). The former denote those layers which take responsibility of processing the structured data by aggregations, and generate vector representations for all the nodes in graph. Task-specific layers are exploited to deal with individual problems (e.g., classification, regression). In Fig. \ref{fig:gnnmol}, task-specific layers are implemented by fully-connected layers which takes the molecular representation as input to output a value. We also point out that the types of layers such as pooling, readout depend on the individual situations. For example, readout-equivalent operation has been implicitly defined within the graph convolution algorithm \cite{duvenaud2015convolutional} because it aims to generate a molecular representations (graph-level), and in this case the readout operation is classified into graph layer. In contrast, graph convolutional networks \cite{kipf2016semi} was originally proposed to learn each node representation for node classification. When it is applied to solve graph-level tasks, the node representations have to be collapsed into a representation by readout, and in this case the readout is classified into task-specific layer. 

Furthermore, there is an increasing number of research works about applying GNN into molecular property prediction. The work presented in \cite{wieder2020compact} taxonomies the problems of molecular property prediction into four main types: quantum chemistry, physical chemistry, biophysics, and biological effect. MPNN~\cite{gilmer2017neural} makes use of the idea of message-passing to update atom hidden states recursively, and the final hidden states of atoms are processed by readout function to predict the multiple properties in quantum chemistry originally. Thereafter, directed MPNN \cite{yang2019analyzing}, MPNN with SELU activation function \cite{withnall2020building}, MPNN with attention and edge memory mechanisms \cite{withnall2020building} have been proposed as the variants of MPNN \cite{gilmer2017neural} and applied to predict a wider range of molecular properties. Message-passing is a kind of propagating information in spatial manner. Similarly, there are some GNNs which have been proposed based on the idea of performing the spatial-like convolution operation on molecular graphs \cite{coley2017convolutional, duvenaud2015convolutional, wang2019molecule, xinyi2018capsule}. In contrast, the number of the applications of spectral GNNs in molecular property prediction is relatively small. LanczosNet \cite{liao2019lanczosnet}, GCN \cite{henaff2015deep}, AGCN \cite{li2018adaptive}, GCN with eigen-pooling
\cite{ma2019graph} which have been proposed based on the spectral decomposition. Furthermore, the work presented in \cite{gilmer2017neural} used MPNN framework to unify message-passing-based, spatial, and spectral GNNs.
% Other operations/layers (e.g., pooling, readout) depend on the situations. For example, there are some GNNs \cite{duvenaud2015convolutional, xiong2019pushing} explicitly/implicitly defined readout operation within their orginal algorithms. Sometimes, readout layer is added by users for solving specific problems.

\subsection{CMA-ES}
CMA-ES\cite{hansen2016cma} is one of the state-of-the-art evolutionary approaches for HPO. It regularly dominates the black-box optimization benchmarking (BBOB) challenge \cite{hutter2019automated}. The core idea of CMA-ES is maintaining a multivariate normal distribution (Eq. 1) to sample new solutions. The multivariate normal distribution consists of a covariance matrix $C$ and mean $m$, and step size $\sigma$. The update of these parameters is inspired by biological evolution. Specifically, $g$ denotes the generation in Eq. 1, CMA-ES always follows the principle of `survival of the fittest', the number of $\mu$ promising individuals are selected to update $m^{(g)}$ with different weights $w$ by Eq. 2. To update $C^{(g)}$, CMA-ES combines the Rank-$\mu$-Update with Rank-One-Update to keep the balance between exploitation and exploration. Finally, to control the scale of step size and speed up learning proccess, CMA-ES introduces $\sigma$ which is updated by cumulative path length control.
\begin{equation}
    \boldsymbol{x}_{k}^{(g+1)} \sim \boldsymbol{m}^{(g)}+\sigma^{(g)} \mathcal{N}\left(\mathbf{0}, \boldsymbol{C}^{(g)}\right) \quad \text { for } k=1, \ldots, \lambda
\end{equation}

\begin{equation}
    \boldsymbol{m}^{(g+1)}=\sum_{i=1}^{\mu} w_{i} \boldsymbol{x}_{i: \lambda}^{(g+1)}
\end{equation}

% \begin{equation}
%     \begin{array}{r}
% \boldsymbol{C}^{(g+1)}=(\underbrace{1-c_{1}-c_{\mu} \sum w_{j}}_{\text {can be close or equal to } 0}) \boldsymbol{C}^{(g)} \\
% \qquad+c_{1} \underbrace{\boldsymbol{p}_{\mathrm{c}}^{(g+1)} \boldsymbol{p}_{\mathrm{c}}^{(g+1)}}_{\text {rank-one update }}+c_{\mu} \underbrace{\sum_{i=1}^{\lambda} w_{i} \boldsymbol{y}_{i: \lambda}^{(g+1)}\left(\boldsymbol{y}_{i: \lambda}^{(g+1)}\right)^{\top}}_{\text {rank- } \mu \text { update }}
% \end{array}
% \end{equation}

The most surprising discovery about CMA-ES is that the learning of the distribution parameters is similar to the descent in direction of a sampled natural gradient of the expected objective function value \cite{akimoto2010bidirectional}. Additionally, the using of convariance matrix in the method provides promising future for dealing with multi-dimensions HPO problems \cite{nomura2020warm}. In concrete, there are some hyperparameters are not independent for all deep learning models, for example, the learning rate and batch size. Convariance matrix helps to automatically build the relation between them and coordinate them smoothly.

\section{Experiments}
ESOL~\cite{delaney2004esol}, FreeSolv~\cite{mobley2014freesolv}, and Lipophilicity~\cite{hersey_chembl_2015} are three representative molecular benchmark datasets used in our experiments, and they respectively correspond the tasks of predicting solubility, hydration free energy, and octanol/water distribution coefficient. The sizes of them are 1,128, 642, 4,200 respectively. Each dataset is split into training, validation, and test sets with the ratio $80\%/10\%/10\%$. The training set is used to train GNNs, the validation set is used for guiding HPO, and test set is used to do final evaluations. Compared with the common designs of HPO experiments, we made the modification that the evaluation of a solution (hyperparameter setting) is repeated three times, and the mean of the root mean squared errors (RMSE) is used to score the solutions.

Meanwhile, we employed the graph convolution model (GC) \cite{duvenaud2015convolutional} to predict these properties in the above mentioned three datasets. GC was proposed with the molecular domain knowledge which fits our research problem (molecular property prediction) compared with other GNNs. For ease of implementation, we leveraged the DeepChem (Python toolkit for deep learning in drug discovery, materials science, quantum chemistry, and biology) \cite{Ramsundar-et-al-2019} to preprocess molecular data and implement GC. In addition, we make use of Optuna \cite{akiba2019optuna} to conduct the HPO experiments with CMA-ES. 

To assess the impact of optimizing different types of hyperparameters on the performance of GNNs, we first look at the graph-related hyperparameters, and for GC, they include the number of graph convolution layer $n_g$, the sizes of the graph convolution layer $s_g$, and the size of dense layer $s_d$ which is defined in GC to generate molecular representations. The range of $n_g$  is $1 \sim 6$ with the step size 1.  $s_g$ and $s_d$ are in the ranges of $32 \sim 512$ (step size 32) and $64 \sim 1024$ (step size 64), respectively. These ranges are set according to the default value provided in \cite{wu2018moleculenet}, while the step sizes are selected following \cite{yuan2021novel}. As for the task-specific hyperparameters (i.e., the hyperparameters in task-specific layers), in order to predict molecular properties $\mathbb{R}$, we employed a simple feedforward neural network which consists of a few fully-connected layers. So the task-specific hyperparameters include the number of fully-connected layers $n_f$ (excluding the output layer), and the sizes of those layers $s_f$, and the activation function $a$. The options of $a$ are $Sigmoid$, $Tanh$, and $ReLU$. Meanwhile, the ranges of $s_f$ referring to $s_d$ are from $64 \sim 1024$ with the step size 64. The arrangement of $n_f$ is the same as $n_g$ for facilitating analysis. The aboved hyperparameters are summarized in Table \ref{tab:hpssumary}. Furthermore, $n_g$ and $n_f$ will determine the number of $s_g$ and $s_f$, so our search space is dynamic. The dynamic feature of the search space will change the dimensions of the problem during the search, which makes HPO more challenging.

\begin{table}[]
\centering
 \resizebox{85mm}{15mm}{
\begin{tabular}{|l|l|l|l|}
\hline
 Types & Hyperparameters & Ranges & Step Sizes \\ \hline
\multirow{3}{*}{Graph Layer} & $n_g$ & $1 \sim 6$ & $1$ \\ \cline{2-4} 
 & $s_g$ & $32 \sim 512$ & $32$ \\ \cline{2-4} 
 & $s_d$ & 64-1024 & $64$ \\ \hline
\multirow{3}{*}{Fully-connected Layer} & $n_f$ & 1-6 & $1$ \\ \cline{2-4} 
 & $s_f$ & $64 \sim 1024$ & $64$ \\ \cline{2-4} 
 & $a$ & $sigmoid, relu, tanh$ & $1$ \\ \hline
\end{tabular}%
}
\caption{Hyperparameters Summary}
\label{tab:hpssumary}
\end{table}

\subsection{Pseudo-dynamic Search Space}
However, it is noted that CMA-ES does not support dynamic search space \cite{akiba2019optuna}. Therefore we turn to implement the pseudo-dynamic process, and the process of HPO is shown in Algorithm \ref{alg1}. In Algorithm \ref{alg1}, $\mathbf{\Lambda}$ denotes the entire hyperparameter space, and $|\mathbf{\Lambda}| = N$ means the number of hyperparameters. Furthermore, $\Lambda_{n}^{I}$ is used to define a dynamic hyperparameter. For example, in our experiments, $s_g$ is a list in which each element represents the size of the corresponding graph convolution layer. Meanwhile, the number of elements in $s_g$ is dynamic as it is determined by $n_g$. It is not possible to decrease/increase the dimensions of the multivariate normal distribution after initialization, so that we keep the CMA-ES to sample the maximum number of elements. Regarding this, we make use of $n_g$ to decide how many elements will be used to instantiate the model. In this way, the search space maintained by CMA-ES is not changed, but in practical it affects the generation of models. 

\begin{algorithm}[!ht]
\SetAlgoLined
\SetKwInOut{Input}{input}
\Input{the hyperparameter space $\mathbf{\Lambda}=\Lambda_{1} \times \Lambda_{2} \times \ldots \Lambda_{N}$, where $\Lambda_{n}$ denotes the $n-$th hyperparameter; if $\Lambda_{n}$ is a dynamic hyperparameter, $\Lambda_{n} = \{ \Lambda_{n}^{1}, \Lambda_{n}^{2}, \ldots \Lambda_{n}^{i}\}$, where $i \in \Lambda_{m}$, $(\Lambda_{m},\Lambda_{n})$ are paired, $\Lambda_{m}$ will determines the number of elements in $\Lambda_{n}$ ; a GNN $\mathcal{M}$; the total number of trials $T$}
 initializing CMA-ES; $\lambda_{best} \text{ current best hyperparameter setting}$\; 
 sort($\Lambda$) \tcp*[r]{move all dynamic hyperparameters backward}
 t = 1\;
 \While{$t < T$}{
 $\lambda$ = [] \tcp*[r]{null list for collecting sampled hyperparameters }
 \For{$n$ = $1$ to $N$}
  {
  \uIf{$\Lambda_{n}$ \text{is a dynamic hyperparameter}}{
    lookup ($(\Lambda_{m},\Lambda_{n})$)\;
    %  $i$ = CMA-ES.$\text{suggest}(\Lambda_{m})$\;
     \For{$z$ = $1$ to $I$}{
        $v$ = CMA-ES.\text{suggest}($\Lambda_{n}^{z}$)\;
        \uIf{$z <= i$}{
           $\lambda$.$\text{append}(v)$\;
        }
     
      }}
        \Else{
        $v$ = CMA-ES.$\text{suggest}(\Lambda_{n})$\;
        $\lambda$.$\text{append}(v)$\;
  }
  }
   $\mathcal{M}$ is instantiated with $\lambda$\;
  evaluate($\mathcal{M}$)\;
    update($\lambda_{best}$)\;
 $t += 1$\;
 }
 \caption{HPO on Pseudo-dynamic Search Space}
 \label{alg1}
\end{algorithm}

Overall, we designed four sets of experiments. The first set of experiments take the default values of hyperparameters from DeepChem \cite{Ramsundar-et-al-2019} to train a GC model for thirty times on ESOL, FreeSolv, and Lipophilicity datasets, respectively. The average of root mean squared errors (Mean RMSE) and standard deviations (Mean Std) are shown in Table \ref{tab:deafault}.  Batch size $s_b$, the number of training epochs $n_e$, and learning rate $l_r$, which are not considered to optimize because this study focuses on discovering the impact of optimizing the two different types of GNNs' hyperparameters. Meanwhile, $s_b = 128$, $n_e = 100$, and $l_r=0.0005$ are set for all our experiments. In Table \ref{tab:deafault}, $n_f$, $s_f$ and $a$ are 0 or none because the task-specific layer only has a single output layer (i.e., no hidden layers) according to the orginal settings in DeepChem \cite{Ramsundar-et-al-2019}. The purpose of this group of experiments is to set a baseline for our following experiments.

\begin{table}[!ht]
\begin{tabular}{|l|l|l|l|l|}
\hline
\multicolumn{2}{|l|}{Datasets}        & ESOL                                                      & FreeSolv                                                    & Lipophilicity                                                    \\ \hline
\multicolumn{2}{|l|}{Hyperparameters} & \multicolumn{3}{l|}{\begin{tabular}[c]{@{}l@{}}$n_g= 2$, $s_g= [128,128]$, \\ $s_d = 256$, $n_f = 0$,\\ $s_f = none$, $a = none$,\\ $s_b = 128$, $n_e = 100$,\\ $l_r = 0.0005$,\\ \end{tabular}} \\ \hline
\multirow{2}{*}{Train}   & Mean RMSE  & 0.3202                                                    & 0.7782                                                      & 0.2146                                                           \\ \cline{2-5} 
                         & Mean Std   & 0.0487                                                    & 0.0785                                                      & 0.0259                                                           \\ \hline
\multirow{2}{*}{Valid}   & Mean RMSE  & 1.1145                                                    & 2.1353                                                      & 0.7245                                                           \\ \cline{2-5} 
                         & Mean Std   & 0.0479                                                    & 0.2326                                                      & 0.0247                                                           \\ \hline
\multirow{2}{*}{Test}    & Mean RMSE  & 1.1570                                                    & 1.8482                                                      & 0.7424                                                           \\ \cline{2-5} 
                         & Mean Std   & 0.0700                                                    & 0.1679                                                      & 0.0203                                                           \\ \hline
\end{tabular}
\caption{The Results by Using the Hyperparameters Provided by MoleculeNet}
\label{tab:deafault}
\end{table}
 
The other three sets of experiments are conducted the HPO for GC on ESOL, FreeSolv, and Lipophilicity, respectively. In Section 3.2, we will discuss the process of HPO on the three datasets (Figs. \ref{fig:esol}$\sim$\ref{fig:lipo}). Thereafter, we will start to analyse the results in details for each dataset (Tables \ref{tab:esol}$\sim$\ref{tab:lipo}).

\subsection{The Process of Hyperparameter Optimization}
\begin{figure*}[!ht]
    \centering
    \includegraphics[scale=0.28]{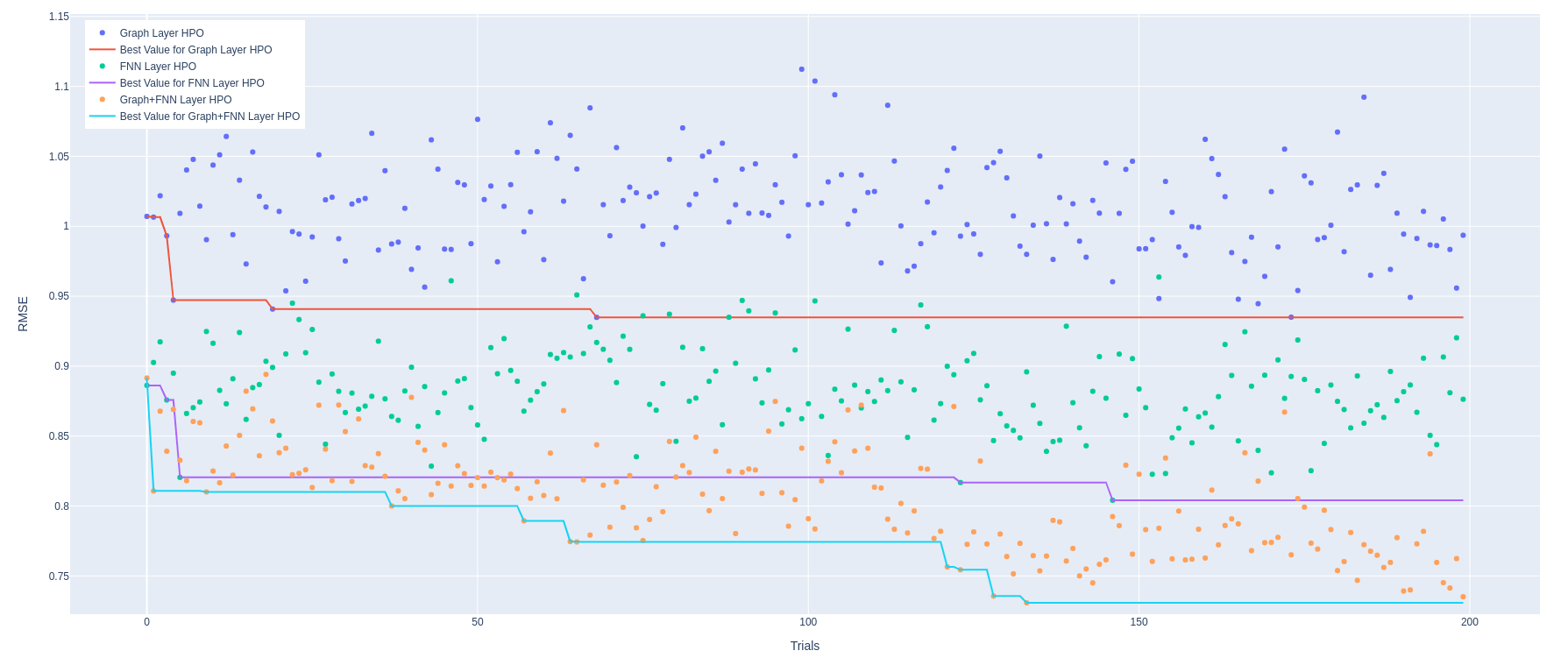}
    \caption{The Optimization Process on the ESOL Dataset}
    \label{fig:esol}
\end{figure*}

\begin{figure*}[!ht]
    \centering
    \includegraphics[scale=0.28]{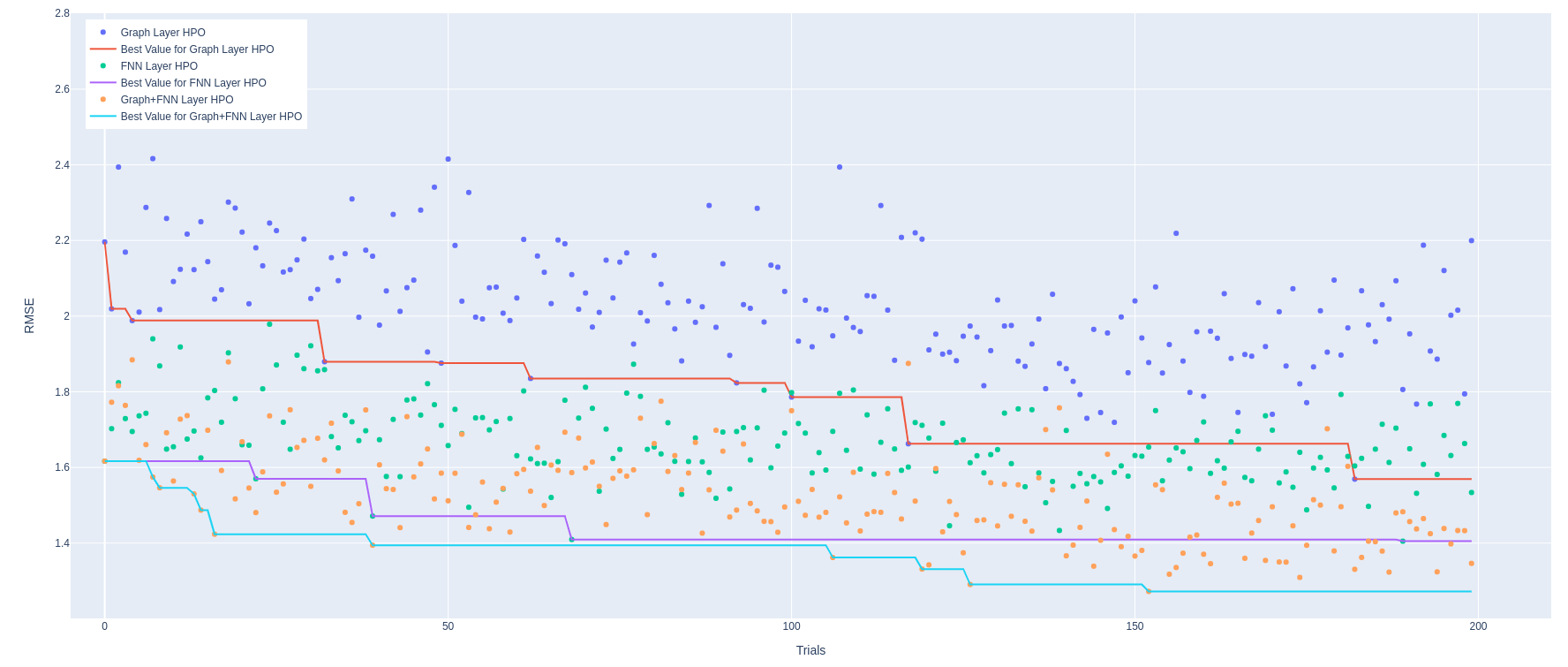}
    \caption{The Optimization Process on the FreeSolv Dataset}
    \label{fig:freesolv}
\end{figure*}

\begin{figure*}[!ht]
    \centering
    \includegraphics[scale=0.28]{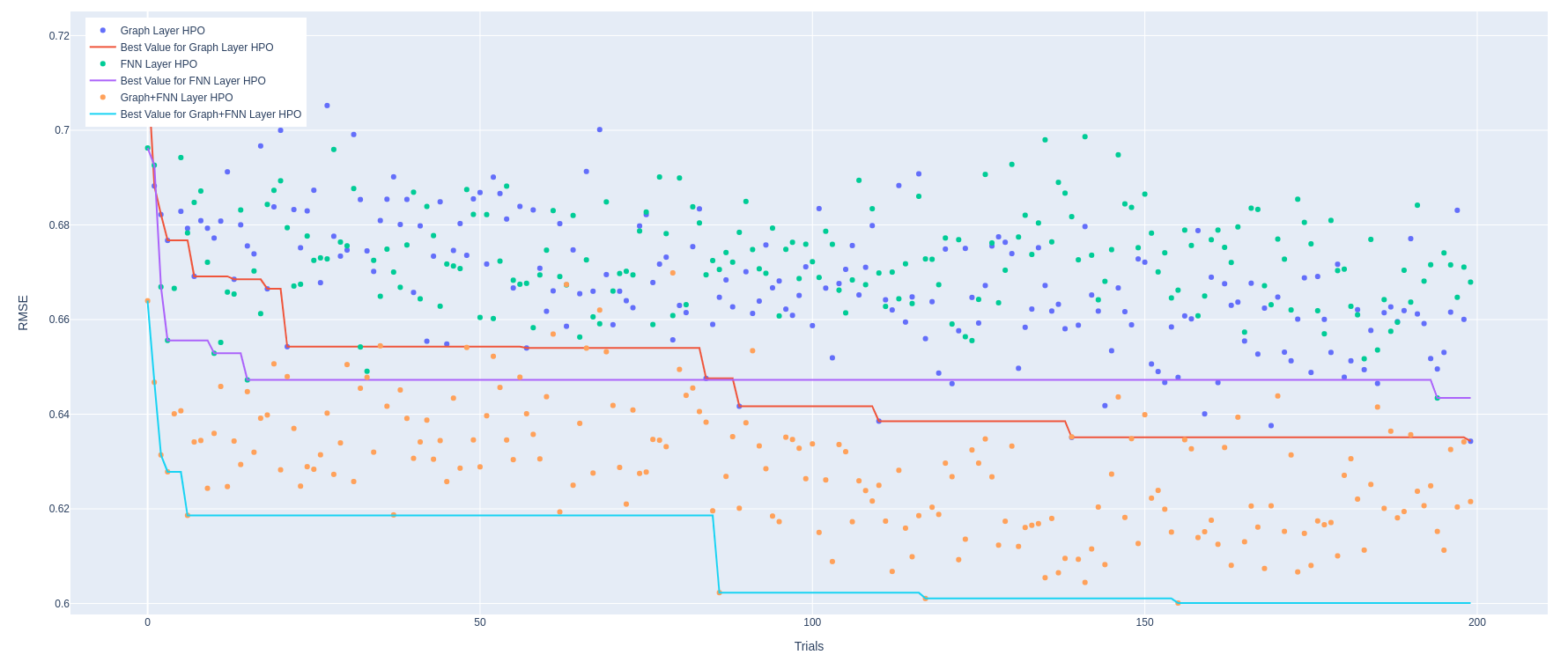}
    \caption{The Optimization Process on the Lipophilicity Dataset}
    \label{fig:lipo}
\end{figure*}

In Figs. \ref{fig:esol}$\sim$\ref{fig:lipo}, the x-axis denotes the index of each trial. One trial represents a process of sampling and evaluating a new hyperparameter setting. CMA-ES was assigned with 200 trials for each dataset. The y-axis denotes the metric of RMSE for evaluating each trial; each trial in our experiments are evaluated thrice, and the mean values are drawn in these figures. The reason for using the mean value of multiple evaluations is that we observed the results for multiple evaluations are not stable, which may mislead the HPO. In Fig. \ref{fig:esol}$\sim$\ref{fig:lipo}, the blue points represent the RMSEs for optimizing graph-related hyperparameters, the green points represent the RMSEs of optimizing task-specified hyperparameters, and the orange points denote the results of optimizing both graph-related and task-specified hyperparameters simultaneously. The nine lines are used to represent the trends of performing HPO on different types of hyperparameters, and these lines are drawn by connecting all current best points in time sequence. It is easy to observe that most of the lines hold obviously decreasing trends, which indicates that CMA-ES works for optimizing hyperparameters. Furthermore, in Figs. \ref{fig:esol} and \ref{fig:lipo}, the decreasing trends of RMSEs for red and purple lines are less significant, compared with those for the blue lines. From these observations, we can see that, Fig. \ref{fig:esol} implies that appropriate settings for both graph and task-specific layers are needed together, and they may complement with each other to achieve better performance in molecular prediction tasks. Overall, the three figures indicate that optimizing both types of hyperparameters can get more gains given the same number of trials, even the search space becomes larger because the number of possible combinations of different hyperparameters increases.

\subsection{The Results}
The best hyperparamter values obtained from Section 3.2 are used to instantiate GCs, and these GCs are trained respectively on ESOL, FreeSolv, and Lipophilicity datasets. The detailed results are shown in Tables \ref{tab:esol} $\sim$\ref{tab:lipo}.

In general, the models configured with the CMA-ES optimized hyperparameters on the three datasets achieved better performances than the original ones (Tab \ref{tab:deafault}). For example, in ESOL, the RMSE of the GC with default hyperparameters (Table \ref{tab:deafault}) is 1.1570 on the test set. The models with HPO on graph layers, task-specific layers, and both of them have the Mean RMSE of 1.0854, 0.9505, and 0.8824, respectively (Table \ref{tab:esol}). To statistically analyse the improvements, we conducted the $t$-test for the RMSEs on the test dataset between Table \ref{tab:deafault} and Table \ref{tab:esol} with the $t$ values of 4.0000, 12.7625, and 18.1311. When the significance level $\alpha = 0.001$, their performance are all significantly better than original ones??

In Tables \ref{tab:esol}$\sim$\ref{tab:lipo}, we can see that HPO on graph and fully connected (task-specific) layers outperforms HPO on either graph or fully connected layers. With the same number of trials in HPO, optimizing graph or fully connected layers face a large search space, but it achieved promising performance. Meanwhile, in Tables \ref{tab:freesolv} and \ref{tab:lipo}, we observed that only optimizing fully-connected layers has relatively more serious over-fitting problem compared with the optimizing both types of hyperparameters, since it always obtained less RMSE values on the training datasets, and inversely had larger RMSE values on validation and test sets. In this case, it indicates optimizing fully-connected layers only would help to fit the problems, but without optimizing graph layers, the molecular representations may not be better learnt, which leads to reduced performance of GNNs in test set. Conducting HPO on graph layers only achieved lower  performance than performing HPO on task-specific layers and the both in the three datasets. We believe the reason is that the default setting of GC only provides a output layer without hidden layers; this means molecular representations are only passed to a linear layer (without non-linear transformation) which dramatically restricts the learning capability. Interestingly, after HPO, the hyperparameter $a$ was assigned to $ReLU$ in all experiments which is the same choice as described in \cite{duvenaud2015convolutional}, where the $ReLU$ activation function was manually selected.

In summary, although graph layers and task-specific layers play different roles in GNNs, they need to be optimized together when solving practical problems. The reason is as follows: a better graph representation learned from graph layers needs to be supported by tailored task-specific layers to accomplish tasks. Similarly, task-specific layers also need appropriate graph representations to achieved good performances.

From the above analysis, we can conclude that when only limited computational budget is available, we should still optimize the hyperparameters of both types of layers, rather than focusing on one of them. 

% Please add the following required packages to your 
\begin{table*}[!ht]
\normalsize
\resizebox{\textwidth}{25mm}{
\begin{tabular}{|l|l|l|l|l|}
\hline
\multicolumn{2}{|l|}{ESOL}                             & Graph Layers                                                                                             & Fully Connected Layers                                                                                                  & Graph and Fully Connected Layers                                                                                                              \\ \hline
\multicolumn{2}{|l|}{\multirow{4}{*}{Hyperparameters}} & \multirow{4}{*}{\begin{tabular}[c]{@{}l@{}}$n_g= 3$, \\ $s_g= [320, 384, 128]$, \\ $s_d = 448$\end{tabular}} & \multirow{4}{*}{\begin{tabular}[c]{@{}l@{}}$n_f = 5$, \\ $s_f = [512, 768, 832, 320, 192]$, \\ $a= relu$\end{tabular}} & \multirow{4}{*}{\begin{tabular}[c]{@{}l@{}}$n_g=1$, $s_g= 512$, \\ $s_d= 640$, $n_f=5$,\\ $s_f= [1024, 832, 640, 192, 832]$,\\ $a = relu$\end{tabular}}  \\
\multicolumn{2}{|l|}{}                                 &                                                                                                          &                                                                                                                         &                                                                                                                                               \\
\multicolumn{2}{|l|}{}                                 &                                                                                                          &                                                                                                                         &                                                                                                                                               \\
\multicolumn{2}{|l|}{}                                 &                                                                                                          &                                                                                                                         &                                                                                                                                               \\ \hline
\multirow{2}{*}{Train}           & Mean RMSE           & 0.3554                                                                                                   & \textbf{0.1917}                                                                                                         & 0.2328                                                                                                                                        \\ \cline{2-5} 
                                 & Mean Std            & 0.0368                                                                                                   & 0.0372                                                                                                                  & 0.0651                                                                                                                                        \\ \hline
\multirow{2}{*}{Valid}           & Mean RMSE           & 1.0108                                                                                                   & 0.8678                                                                                                                  & \textbf{0.7718}                                                                                                                               \\ \cline{2-5} 
                                 & Mean Std            & 0.0554                                                                                                   & 0.0318                                                                                                                  & 0.0371                                                                                                                                        \\ \hline
\multirow{2}{*}{Test}            & Mean RMSE           & 1.0854                                                                                                   & 0.9508                                                                                                                  & \textbf{0.8824}                                                                                                                               \\ \cline{2-5} 
                                 & Mean Std            & 0.0660                                                                                                   & 0.0515                                                                                                                  & 0.0417                                                                                                                                        \\ \hline
\end{tabular}}
\caption{HPO on the ESOL Dataset}
\label{tab:esol}
\end{table*}

\begin{table*}[!ht]
\normalsize
\resizebox{\textwidth}{25mm}{
\begin{tabular}{|l|l|l|l|l|}
\hline
\multicolumn{2}{|l|}{FreeSolv}                         & Graph Layers                                                                                   & Fully Connected Layers                                                                               & Graph and Fully Connected Layers                                                                                                                  \\ \hline
\multicolumn{2}{|l|}{\multirow{4}{*}{Hyperparameters}} & \multirow{4}{*}{\begin{tabular}[c]{@{}l@{}}$n_g= 1$, \\ $s_g= [256]$, \\ $s_d = 192$\end{tabular}} & \multirow{4}{*}{\begin{tabular}[c]{@{}l@{}}$n_f = 2$, \\ $s_f = [1024, 448]$, \\ $a = relu$\end{tabular}} & \multirow{4}{*}{\begin{tabular}[c]{@{}l@{}}$n_g=2$, $s_g= [512, 352]$,\\  $s_d= 128$, $n_f=5$,\\ $s_f= [192, 640, 320, 320, 768]$, \\$a = relu$\end{tabular}} \\
\multicolumn{2}{|l|}{}                                 &                                                                                                &                                                                                                      &                                                                                                                                                   \\
\multicolumn{2}{|l|}{}                                 &                                                                                                &                                                                                                      &                                                                                                                                                   \\
\multicolumn{2}{|l|}{}                                 &                                                                                                &                                                                                                      &                                                                                                                                                   \\ \hline
\multirow{2}{*}{Train}           & Mean RMSE           & 0.7743                                                                                         & \textbf{0.4565}                                                                                      & 0.4926                                                                                                                                            \\ \cline{2-5} 
                                 & Mean Std            & 0.1361                                                                                         & 0.2301                                                                                               & 0.1510                                                                                                                                            \\ \hline
\multirow{2}{*}{Valid}           & Mean RMSE           & 1.9128                                                                                         & 1.6702                                                                                               & \textbf{1.4779}                                                                                                                                   \\ \cline{2-5} 
                                 & Mean Std            & 0.2172                                                                                         & 0.1464                                                                                               & 0.1191                                                                                                                                            \\ \hline
\multirow{2}{*}{Test}            & Mean RMSE           & 1.6983                                                                                         & 1.4118                                                                                               & \textbf{1.2285}                                                                                                                                   \\ \cline{2-5} 
                                 & Mean Std            & 0.1891                                                                                         & 0.1461                                                                                               & 0.0727                                                                                                                                            \\ \hline
\end{tabular}}
\caption{HPO on the FreeSolv Dataset}
\label{tab:freesolv}
\end{table*}

\begin{table*}[!ht]
\normalsize
\resizebox{\textwidth}{26mm}{
\begin{tabular}{|l|l|l|l|l|}
\hline
\multicolumn{2}{|l|}{Lipophilicity}                    & Graph Layers                                                                                                           & Fully Connected Layers                                                                                       & Graph and Fully Connected Layers                                                                                                                           \\ \hline
\multicolumn{2}{|l|}{\multirow{4}{*}{Hyperparameters}} & \multirow{4}{*}{\begin{tabular}[c]{@{}l@{}}$n_g= 6$,\\ $s_g= [416, 256, 512, 320, 384, 128]$, \\ $s_d = 768$\end{tabular}} & \multirow{4}{*}{\begin{tabular}[c]{@{}l@{}}$n_f = 4$,\\ $s_f = [1024, 896, 832, 64]$, \\ $a= relu$\end{tabular}} & \multirow{4}{*}{\begin{tabular}[c]{@{}l@{}}$n_g=5$, $s_g= [480, 512, 256, 192, 224]$,\\ $s_d= 960$, $n_f=4$,\\ $s_f= [704, 320, 128, 768]$,\\ $a = relu$\end{tabular}} \\
\multicolumn{2}{|l|}{}                                 &                                                                                                                        &                                                                                                              &                                                                                                                                                            \\
\multicolumn{2}{|l|}{}                                 &                                                                                                                        &                                                                                                              &                                                                                                                                                            \\
\multicolumn{2}{|l|}{}                                 &                                                                                                                        &                                                                                                              &                                                                                                                                                            \\ \hline
\multirow{2}{*}{Train}            & Mean RMSE          & 0.2148                                                                                                                 & \textbf{0.1369}                                                                                              & 0.1701                                                                                                                                                     \\ \cline{2-5} 
                                  & Mean Std           & 0.0206                                                                                                                 & 0.0201                                                                                                       & 0.0361                                                                                                                                                     \\ \hline
\multirow{2}{*}{Valid}            & Mean RMSE          & 0.6655                                                                                                                 & 0.6656                                                                                                       & \textbf{0.6239}                                                                                                                                            \\ \cline{2-5} 
                                  & Mean Std           & 0.0171                                                                                                                 & 0.0144                                                                                                       & 0.0154                                                                                                                                                     \\ \hline
\multirow{2}{*}{Test}             & Mean RMSE          & 0.7014                                                                                                                 & 0.6786                                                                                                       & \textbf{0.6472}                                                                                                                                            \\ \cline{2-5} 
                                  & Mean Std           & 0.0148                                                                                                                 & 0.0165                                                                                                       & 0.0187                                                                                                                                                     \\ \hline
\end{tabular}}
\caption{HPO on the Lipophilicity Dataset }
\label{tab:lipo}
\end{table*}

\section{Conclusions and Future Work}
With the rapid development of GNNs, applying them in molecular machine learning problems becomes increasingly compelling and meaningful. For example, accurate molecular property prediction can significantly facilitate the entire process of drug discovery in a faster and cheaper way. However, the performance of GNNs are largely affected by hyperparameter selection, so the research of HPO on GNNs is of extremely important. 

In this paper, we elaborated the problem of HPO on GNNs for molecular property prediction, and investigated in depth that which types of hyperparameters should be selected to optimize when computational resources are limited. Based on our experiments, we conclude that both graph-related hyperparameters and task-specific hyperparameters should be optimised simultaneously, and leaving any one out will result in reduced performance. Even doing this means a larger search space, which seems to be more challenging given the same number of trials (limited computational resources), such a strategy can surprisingly achieve better performance.

Finally, we acknowledge that our experiments are based on one type of GNN model and one evolutionary strategy. However, we believe that our conclusion can be further generalised, because we have selected the representative GNN model, used state-of-the-art evolutionary HPO approach, and the benchmark datasets used for experiments are also representative in molecular property prediction problems. 

Still, we propose two future directions to carry out our research in the next step. First, there exist various GNNs, and most of them comply with the rule of aggregating the neighbor information to learn the node representations. However, they can be classified into spectral and spatial GNNs \cite{balcilar2020bridging}. In this research, we have extensively investigated the impact of HPO on GC, which is a representative of spatial GNNs. Therefore, it would be interesting and worthwhile to investigate whether the same conclusion holds for HPO of spectral GNNs. Second, we employed CMA-ES as the HPO strategy because it is a state-of-the-art evolutionary HPO method. However, it does not support the dynamic search space, which constrains its scalability. In our future work, other evolutionary HPO approaches can be applied to explore their effectiveness on optimizing hyperparamters with dynamic search space for GNNs.
\bibliographystyle{ACM-Reference-Format}
\bibliography{sample-base}


\begin{thebibliography}{41}

%%% the Web address.

\ifx \showCODEN    \undefined \def \showCODEN     #1{\unskip}     \fi
\ifx \showDOI      \undefined \def \showDOI       #1{#1}\fi
\ifx \showISBNx    \undefined \def \showISBNx     #1{\unskip}     \fi
\ifx \showISBNxiii \undefined \def \showISBNxiii  #1{\unskip}     \fi
\ifx \showISSN     \undefined \def \showISSN      #1{\unskip}     \fi
\ifx \showLCCN     \undefined \def \showLCCN      #1{\unskip}     \fi
\ifx \shownote     \undefined \def \shownote      #1{#1}          \fi
\ifx \showarticletitle \undefined \def \showarticletitle #1{#1}   \fi
\ifx \showURL      \undefined \def \showURL       {\relax}        \fi
% invisible to TeX
\providecommand\bibfield[2]{#2}
\providecommand\bibinfo[2]{#2}
\providecommand\natexlab[1]{#1}
\providecommand\showeprint[2][]{arXiv:#2}

\bibitem[\protect\citeauthoryear{Akiba, Sano, Yanase, Ohta, and Koyama}{Akiba
  et~al\mbox{.}}{2019}]%
        {akiba2019optuna}
\bibfield{author}{\bibinfo{person}{Takuya Akiba}, \bibinfo{person}{Shotaro
  Sano}, \bibinfo{person}{Toshihiko Yanase}, \bibinfo{person}{Takeru Ohta},
  {and} \bibinfo{person}{Masanori Koyama}.} \bibinfo{year}{2019}\natexlab{}.
\newblock \showarticletitle{Optuna: A next-generation hyperparameter
  optimization framework}. In \bibinfo{booktitle}{\emph{Proceedings of the 25th
  ACM SIGKDD international conference on knowledge discovery \& data mining}}.
  \bibinfo{pages}{2623--2631}.
\newblock


\bibitem[\protect\citeauthoryear{Akimoto, Nagata, Ono, and Kobayashi}{Akimoto
  et~al\mbox{.}}{2010}]%
        {akimoto2010bidirectional}
\bibfield{author}{\bibinfo{person}{Youhei Akimoto}, \bibinfo{person}{Yuichi
  Nagata}, \bibinfo{person}{Isao Ono}, {and} \bibinfo{person}{Shigenobu
  Kobayashi}.} \bibinfo{year}{2010}\natexlab{}.
\newblock \showarticletitle{Bidirectional relation between CMA evolution
  strategies and natural evolution strategies}. In
  \bibinfo{booktitle}{\emph{International Conference on Parallel Problem
  Solving from Nature}}. Springer, \bibinfo{pages}{154--163}.
\newblock


\bibitem[\protect\citeauthoryear{Atwood and Towsley}{Atwood and
  Towsley}{2016}]%
        {atwood2016diffusion}
\bibfield{author}{\bibinfo{person}{James Atwood} {and} \bibinfo{person}{Don
  Towsley}.} \bibinfo{year}{2016}\natexlab{}.
\newblock \showarticletitle{Diffusion-convolutional neural networks}. In
  \bibinfo{booktitle}{\emph{Advances in neural information processing
  systems}}. \bibinfo{pages}{1993--2001}.
\newblock


\bibitem[\protect\citeauthoryear{Balcilar, Renton, H{\'e}roux, Gauzere, Adam,
  and Honeine}{Balcilar et~al\mbox{.}}{2020}]%
        {balcilar2020bridging}
\bibfield{author}{\bibinfo{person}{Muhammet Balcilar},
  \bibinfo{person}{Guillaume Renton}, \bibinfo{person}{Pierre H{\'e}roux},
  \bibinfo{person}{Benoit Gauzere}, \bibinfo{person}{Sebastien Adam}, {and}
  \bibinfo{person}{Paul Honeine}.} \bibinfo{year}{2020}\natexlab{}.
\newblock \showarticletitle{Bridging the gap between spectral and spatial
  domains in graph neural networks}.
\newblock \bibinfo{journal}{\emph{arXiv preprint arXiv:2003.11702}}
  (\bibinfo{year}{2020}).
\newblock


\bibitem[\protect\citeauthoryear{Bergstra, Bardenet, Bengio, and
  K{\'e}gl}{Bergstra et~al\mbox{.}}{2011}]%
        {bergstra2011algorithms}
\bibfield{author}{\bibinfo{person}{James Bergstra}, \bibinfo{person}{R{\'e}mi
  Bardenet}, \bibinfo{person}{Yoshua Bengio}, {and} \bibinfo{person}{Bal{\'a}zs
  K{\'e}gl}.} \bibinfo{year}{2011}\natexlab{}.
\newblock \showarticletitle{Algorithms for hyper-parameter optimization}. In
  \bibinfo{booktitle}{\emph{25th annual conference on neural information
  processing systems (NIPS 2011)}}, Vol.~\bibinfo{volume}{24}. Neural
  Information Processing Systems Foundation.
\newblock


\bibitem[\protect\citeauthoryear{Coley, Barzilay, Green, Jaakkola, and
  Jensen}{Coley et~al\mbox{.}}{2017}]%
        {coley2017convolutional}
\bibfield{author}{\bibinfo{person}{Connor~W Coley}, \bibinfo{person}{Regina
  Barzilay}, \bibinfo{person}{William~H Green}, \bibinfo{person}{Tommi~S
  Jaakkola}, {and} \bibinfo{person}{Klavs~F Jensen}.}
  \bibinfo{year}{2017}\natexlab{}.
\newblock \showarticletitle{Convolutional embedding of attributed molecular
  graphs for physical property prediction}.
\newblock \bibinfo{journal}{\emph{Journal of chemical information and
  modeling}} \bibinfo{volume}{57}, \bibinfo{number}{8} (\bibinfo{year}{2017}),
  \bibinfo{pages}{1757--1772}.
\newblock


\bibitem[\protect\citeauthoryear{Cummings and Nassar}{Cummings and
  Nassar}{2020}]%
        {cummings2020structured}
\bibfield{author}{\bibinfo{person}{Daniel Cummings} {and}
  \bibinfo{person}{Marcel Nassar}.} \bibinfo{year}{2020}\natexlab{}.
\newblock \showarticletitle{Structured Citation Trend Prediction Using Graph
  Neural Networks}. In \bibinfo{booktitle}{\emph{ICASSP 2020-2020 IEEE
  International Conference on Acoustics, Speech and Signal Processing
  (ICASSP)}}. IEEE, \bibinfo{pages}{3897--3901}.
\newblock


\bibitem[\protect\citeauthoryear{Delaney}{Delaney}{2004}]%
        {delaney2004esol}
\bibfield{author}{\bibinfo{person}{John~S Delaney}.}
  \bibinfo{year}{2004}\natexlab{}.
\newblock \showarticletitle{ESOL: estimating aqueous solubility directly from
  molecular structure}.
\newblock \bibinfo{journal}{\emph{Journal of chemical information and computer
  sciences}} \bibinfo{volume}{44}, \bibinfo{number}{3} (\bibinfo{year}{2004}),
  \bibinfo{pages}{1000--1005}.
\newblock


\bibitem[\protect\citeauthoryear{Duvenaud, Maclaurin, Aguilera-Iparraguirre,
  G{\'o}mez-Bombarelli, Hirzel, Aspuru-Guzik, and Adams}{Duvenaud
  et~al\mbox{.}}{2015}]%
        {duvenaud2015convolutional}
\bibfield{author}{\bibinfo{person}{David Duvenaud}, \bibinfo{person}{Dougal
  Maclaurin}, \bibinfo{person}{Jorge Aguilera-Iparraguirre},
  \bibinfo{person}{Rafael G{\'o}mez-Bombarelli}, \bibinfo{person}{Timothy
  Hirzel}, \bibinfo{person}{Al{\'a}n Aspuru-Guzik}, {and}
  \bibinfo{person}{Ryan~P Adams}.} \bibinfo{year}{2015}\natexlab{}.
\newblock \showarticletitle{Convolutional networks on graphs for learning
  molecular fingerprints}.
\newblock \bibinfo{journal}{\emph{arXiv preprint arXiv:1509.09292}}
  (\bibinfo{year}{2015}).
\newblock


\bibitem[\protect\citeauthoryear{Fan, Ma, Li, He, Zhao, Tang, and Yin}{Fan
  et~al\mbox{.}}{2019}]%
        {fan2019graph}
\bibfield{author}{\bibinfo{person}{Wenqi Fan}, \bibinfo{person}{Yao Ma},
  \bibinfo{person}{Qing Li}, \bibinfo{person}{Yuan He}, \bibinfo{person}{Eric
  Zhao}, \bibinfo{person}{Jiliang Tang}, {and} \bibinfo{person}{Dawei Yin}.}
  \bibinfo{year}{2019}\natexlab{}.
\newblock \showarticletitle{Graph neural networks for social recommendation}.
  In \bibinfo{booktitle}{\emph{The World Wide Web Conference}}.
  \bibinfo{pages}{417--426}.
\newblock


\bibitem[\protect\citeauthoryear{Gilmer, Schoenholz, Riley, Vinyals, and
  Dahl}{Gilmer et~al\mbox{.}}{2017}]%
        {gilmer2017neural}
\bibfield{author}{\bibinfo{person}{Justin Gilmer}, \bibinfo{person}{Samuel~S
  Schoenholz}, \bibinfo{person}{Patrick~F Riley}, \bibinfo{person}{Oriol
  Vinyals}, {and} \bibinfo{person}{George~E Dahl}.}
  \bibinfo{year}{2017}\natexlab{}.
\newblock \showarticletitle{Neural message passing for quantum chemistry}. In
  \bibinfo{booktitle}{\emph{International Conference on Machine Learning}}.
  PMLR, \bibinfo{pages}{1263--1272}.
\newblock


\bibitem[\protect\citeauthoryear{Glen, Bender, Arnby, Carlsson, Boyer, and
  Smith}{Glen et~al\mbox{.}}{2006}]%
        {glen2006circular}
\bibfield{author}{\bibinfo{person}{Robert~C Glen}, \bibinfo{person}{Andreas
  Bender}, \bibinfo{person}{Catrin~H Arnby}, \bibinfo{person}{Lars Carlsson},
  \bibinfo{person}{Scott Boyer}, {and} \bibinfo{person}{James Smith}.}
  \bibinfo{year}{2006}\natexlab{}.
\newblock \showarticletitle{Circular fingerprints: flexible molecular
  descriptors with applications from physical chemistry to ADME}.
\newblock \bibinfo{journal}{\emph{IDrugs}} \bibinfo{volume}{9},
  \bibinfo{number}{3} (\bibinfo{year}{2006}), \bibinfo{pages}{199}.
\newblock


\bibitem[\protect\citeauthoryear{Gori, Monfardini, and Scarselli}{Gori
  et~al\mbox{.}}{2005}]%
        {gori2005new}
\bibfield{author}{\bibinfo{person}{Marco Gori}, \bibinfo{person}{Gabriele
  Monfardini}, {and} \bibinfo{person}{Franco Scarselli}.}
  \bibinfo{year}{2005}\natexlab{}.
\newblock \showarticletitle{A new model for learning in graph domains}. In
  \bibinfo{booktitle}{\emph{Proceedings. 2005 IEEE International Joint
  Conference on Neural Networks, 2005.}}, Vol.~\bibinfo{volume}{2}. IEEE,
  \bibinfo{pages}{729--734}.
\newblock


\bibitem[\protect\citeauthoryear{Hansen}{Hansen}{2016}]%
        {hansen2016cma}
\bibfield{author}{\bibinfo{person}{Nikolaus Hansen}.}
  \bibinfo{year}{2016}\natexlab{}.
\newblock \showarticletitle{The CMA evolution strategy: A tutorial}.
\newblock \bibinfo{journal}{\emph{arXiv preprint arXiv:1604.00772}}
  (\bibinfo{year}{2016}).
\newblock


\bibitem[\protect\citeauthoryear{Henaff, Bruna, and LeCun}{Henaff
  et~al\mbox{.}}{2015}]%
        {henaff2015deep}
\bibfield{author}{\bibinfo{person}{Mikael Henaff}, \bibinfo{person}{Joan
  Bruna}, {and} \bibinfo{person}{Yann LeCun}.} \bibinfo{year}{2015}\natexlab{}.
\newblock \showarticletitle{Deep convolutional networks on graph-structured
  data}.
\newblock \bibinfo{journal}{\emph{arXiv preprint arXiv:1506.05163}}
  (\bibinfo{year}{2015}).
\newblock


\bibitem[\protect\citeauthoryear{Hersey}{Hersey}{2015}]%
        {hersey_chembl_2015}
\bibfield{author}{\bibinfo{person}{Anne Hersey}.}
  \bibinfo{year}{2015}\natexlab{}.
\newblock \bibinfo{booktitle}{\emph{{ChEMBL} {Deposited} {Data} {Set} -
  {AZ}\_dataset}}.
\newblock \bibinfo{type}{{T}echnical {R}eport}.
  \bibinfo{institution}{EMBL-EBI}.
\newblock
\urldef\tempurl%
\url{https://doi.org/10.6019/CHEMBL3301361}
\showDOI{\tempurl}


\bibitem[\protect\citeauthoryear{Hutter, Kotthoff, and Vanschoren}{Hutter
  et~al\mbox{.}}{2019}]%
        {hutter2019automated}
\bibfield{author}{\bibinfo{person}{Frank Hutter}, \bibinfo{person}{Lars
  Kotthoff}, {and} \bibinfo{person}{Joaquin Vanschoren}.}
  \bibinfo{year}{2019}\natexlab{}.
\newblock \bibinfo{booktitle}{\emph{Automated machine learning: methods,
  systems, challenges}}.
\newblock \bibinfo{publisher}{Springer Nature}.
\newblock


\bibitem[\protect\citeauthoryear{Jamieson and Talwalkar}{Jamieson and
  Talwalkar}{2016}]%
        {jamieson2016non}
\bibfield{author}{\bibinfo{person}{Kevin Jamieson} {and} \bibinfo{person}{Ameet
  Talwalkar}.} \bibinfo{year}{2016}\natexlab{}.
\newblock \showarticletitle{Non-stochastic best arm identification and
  hyperparameter optimization}. In \bibinfo{booktitle}{\emph{Artificial
  Intelligence and Statistics}}. PMLR, \bibinfo{pages}{240--248}.
\newblock


\bibitem[\protect\citeauthoryear{Jiang, Wu, Hsieh, Chen, Liao, Wang, Shen, Cao,
  Wu, and Hou}{Jiang et~al\mbox{.}}{2021}]%
        {jiang2021could}
\bibfield{author}{\bibinfo{person}{Dejun Jiang}, \bibinfo{person}{Zhenxing Wu},
  \bibinfo{person}{Chang-Yu Hsieh}, \bibinfo{person}{Guangyong Chen},
  \bibinfo{person}{Ben Liao}, \bibinfo{person}{Zhe Wang}, \bibinfo{person}{Chao
  Shen}, \bibinfo{person}{Dongsheng Cao}, \bibinfo{person}{Jian Wu}, {and}
  \bibinfo{person}{Tingjun Hou}.} \bibinfo{year}{2021}\natexlab{}.
\newblock \showarticletitle{Could graph neural networks learn better molecular
  representation for drug discovery? A comparison study of descriptor-based and
  graph-based models}.
\newblock \bibinfo{journal}{\emph{Journal of cheminformatics}}
  \bibinfo{volume}{13}, \bibinfo{number}{1} (\bibinfo{year}{2021}),
  \bibinfo{pages}{1--23}.
\newblock


\bibitem[\protect\citeauthoryear{Karnin, Koren, and Somekh}{Karnin
  et~al\mbox{.}}{2013}]%
        {karnin2013almost}
\bibfield{author}{\bibinfo{person}{Zohar Karnin}, \bibinfo{person}{Tomer
  Koren}, {and} \bibinfo{person}{Oren Somekh}.}
  \bibinfo{year}{2013}\natexlab{}.
\newblock \showarticletitle{Almost optimal exploration in multi-armed bandits}.
  In \bibinfo{booktitle}{\emph{International Conference on Machine Learning}}.
  PMLR, \bibinfo{pages}{1238--1246}.
\newblock


\bibitem[\protect\citeauthoryear{Kipf and Welling}{Kipf and Welling}{2016}]%
        {kipf2016semi}
\bibfield{author}{\bibinfo{person}{Thomas~N Kipf} {and} \bibinfo{person}{Max
  Welling}.} \bibinfo{year}{2016}\natexlab{}.
\newblock \showarticletitle{Semi-supervised classification with graph
  convolutional networks}.
\newblock \bibinfo{journal}{\emph{arXiv preprint arXiv:1609.02907}}
  (\bibinfo{year}{2016}).
\newblock


\bibitem[\protect\citeauthoryear{Li, Wang, Zhu, and Huang}{Li
  et~al\mbox{.}}{2018}]%
        {li2018adaptive}
\bibfield{author}{\bibinfo{person}{Ruoyu Li}, \bibinfo{person}{Sheng Wang},
  \bibinfo{person}{Feiyun Zhu}, {and} \bibinfo{person}{Junzhou Huang}.}
  \bibinfo{year}{2018}\natexlab{}.
\newblock \showarticletitle{Adaptive graph convolutional neural networks}. In
  \bibinfo{booktitle}{\emph{Proceedings of the AAAI Conference on Artificial
  Intelligence}}, Vol.~\bibinfo{volume}{32}.
\newblock


\bibitem[\protect\citeauthoryear{Li, Tarlow, Brockschmidt, and Zemel}{Li
  et~al\mbox{.}}{2015}]%
        {li2015gated}
\bibfield{author}{\bibinfo{person}{Yujia Li}, \bibinfo{person}{Daniel Tarlow},
  \bibinfo{person}{Marc Brockschmidt}, {and} \bibinfo{person}{Richard Zemel}.}
  \bibinfo{year}{2015}\natexlab{}.
\newblock \showarticletitle{Gated graph sequence neural networks}.
\newblock \bibinfo{journal}{\emph{arXiv preprint arXiv:1511.05493}}
  (\bibinfo{year}{2015}).
\newblock


\bibitem[\protect\citeauthoryear{Liao, Zhao, Urtasun, and Zemel}{Liao
  et~al\mbox{.}}{2019}]%
        {liao2019lanczosnet}
\bibfield{author}{\bibinfo{person}{Renjie Liao}, \bibinfo{person}{Zhizhen
  Zhao}, \bibinfo{person}{Raquel Urtasun}, {and} \bibinfo{person}{Richard~S
  Zemel}.} \bibinfo{year}{2019}\natexlab{}.
\newblock \showarticletitle{Lanczosnet: Multi-scale deep graph convolutional
  networks}.
\newblock \bibinfo{journal}{\emph{arXiv preprint arXiv:1901.01484}}
  (\bibinfo{year}{2019}).
\newblock


\bibitem[\protect\citeauthoryear{Ma, Wang, Aggarwal, and Tang}{Ma
  et~al\mbox{.}}{2019}]%
        {ma2019graph}
\bibfield{author}{\bibinfo{person}{Yao Ma}, \bibinfo{person}{Suhang Wang},
  \bibinfo{person}{Charu~C Aggarwal}, {and} \bibinfo{person}{Jiliang Tang}.}
  \bibinfo{year}{2019}\natexlab{}.
\newblock \showarticletitle{Graph convolutional networks with eigenpooling}. In
  \bibinfo{booktitle}{\emph{Proceedings of the 25th ACM SIGKDD International
  Conference on Knowledge Discovery \& Data Mining}}.
  \bibinfo{pages}{723--731}.
\newblock


\bibitem[\protect\citeauthoryear{Mobley and Guthrie}{Mobley and
  Guthrie}{2014}]%
        {mobley2014freesolv}
\bibfield{author}{\bibinfo{person}{David~L Mobley} {and}
  \bibinfo{person}{J~Peter Guthrie}.} \bibinfo{year}{2014}\natexlab{}.
\newblock \showarticletitle{FreeSolv: a database of experimental and calculated
  hydration free energies, with input files}.
\newblock \bibinfo{journal}{\emph{Journal of computer-aided molecular design}}
  \bibinfo{volume}{28}, \bibinfo{number}{7} (\bibinfo{year}{2014}),
  \bibinfo{pages}{711--720}.
\newblock


\bibitem[\protect\citeauthoryear{Nomura, Watanabe, Akimoto, Ozaki, and
  Onishi}{Nomura et~al\mbox{.}}{2020}]%
        {nomura2020warm}
\bibfield{author}{\bibinfo{person}{Masahiro Nomura}, \bibinfo{person}{Shuhei
  Watanabe}, \bibinfo{person}{Youhei Akimoto}, \bibinfo{person}{Yoshihiko
  Ozaki}, {and} \bibinfo{person}{Masaki Onishi}.}
  \bibinfo{year}{2020}\natexlab{}.
\newblock \showarticletitle{Warm Starting CMA-ES for Hyperparameter
  Optimization}.
\newblock \bibinfo{journal}{\emph{arXiv preprint arXiv:2012.06932}}
  (\bibinfo{year}{2020}).
\newblock


\bibitem[\protect\citeauthoryear{Ramsundar, Eastman, Walters, Pande, Leswing,
  and Wu}{Ramsundar et~al\mbox{.}}{2019}]%
        {Ramsundar-et-al-2019}
\bibfield{author}{\bibinfo{person}{Bharath Ramsundar}, \bibinfo{person}{Peter
  Eastman}, \bibinfo{person}{Patrick Walters}, \bibinfo{person}{Vijay Pande},
  \bibinfo{person}{Karl Leswing}, {and} \bibinfo{person}{Zhenqin Wu}.}
  \bibinfo{year}{2019}\natexlab{}.
\newblock \bibinfo{booktitle}{\emph{Deep Learning for the Life Sciences}}.
\newblock \bibinfo{publisher}{O'Reilly Media}.
\newblock
\newblock
\shownote{\url{https://www.amazon.com/Deep-Learning-Life-Sciences-Microscopy/dp/1492039837}.}


\bibitem[\protect\citeauthoryear{Wang, Li, Jiang, Wang, Zhang, and Wei}{Wang
  et~al\mbox{.}}{2019}]%
        {wang2019molecule}
\bibfield{author}{\bibinfo{person}{Xiaofeng Wang}, \bibinfo{person}{Zhen Li},
  \bibinfo{person}{Mingjian Jiang}, \bibinfo{person}{Shuang Wang},
  \bibinfo{person}{Shugang Zhang}, {and} \bibinfo{person}{Zhiqiang Wei}.}
  \bibinfo{year}{2019}\natexlab{}.
\newblock \showarticletitle{Molecule property prediction based on spatial graph
  embedding}.
\newblock \bibinfo{journal}{\emph{Journal of chemical information and
  modeling}} \bibinfo{volume}{59}, \bibinfo{number}{9} (\bibinfo{year}{2019}),
  \bibinfo{pages}{3817--3828}.
\newblock


\bibitem[\protect\citeauthoryear{Weininger}{Weininger}{1988}]%
        {weininger1988smiles}
\bibfield{author}{\bibinfo{person}{David Weininger}.}
  \bibinfo{year}{1988}\natexlab{}.
\newblock \showarticletitle{SMILES, a chemical language and information system.
  1. Introduction to methodology and encoding rules}.
\newblock \bibinfo{journal}{\emph{Journal of chemical information and computer
  sciences}} \bibinfo{volume}{28}, \bibinfo{number}{1} (\bibinfo{year}{1988}),
  \bibinfo{pages}{31--36}.
\newblock


\bibitem[\protect\citeauthoryear{Wieder, Kohlbacher, Kuenemann, Garon, Ducrot,
  Seidel, and Langer}{Wieder et~al\mbox{.}}{2020}]%
        {wieder2020compact}
\bibfield{author}{\bibinfo{person}{Oliver Wieder}, \bibinfo{person}{Stefan
  Kohlbacher}, \bibinfo{person}{M{\'e}laine Kuenemann}, \bibinfo{person}{Arthur
  Garon}, \bibinfo{person}{Pierre Ducrot}, \bibinfo{person}{Thomas Seidel},
  {and} \bibinfo{person}{Thierry Langer}.} \bibinfo{year}{2020}\natexlab{}.
\newblock \showarticletitle{A compact review of molecular property prediction
  with graph neural networks}.
\newblock \bibinfo{journal}{\emph{Drug Discovery Today: Technologies}}
  (\bibinfo{year}{2020}).
\newblock


\bibitem[\protect\citeauthoryear{Withnall, Lindel{\"o}f, Engkvist, and
  Chen}{Withnall et~al\mbox{.}}{2020}]%
        {withnall2020building}
\bibfield{author}{\bibinfo{person}{Michael Withnall}, \bibinfo{person}{Edvard
  Lindel{\"o}f}, \bibinfo{person}{Ola Engkvist}, {and}
  \bibinfo{person}{Hongming Chen}.} \bibinfo{year}{2020}\natexlab{}.
\newblock \showarticletitle{Building attention and edge message passing neural
  networks for bioactivity and physical--chemical property prediction}.
\newblock \bibinfo{journal}{\emph{Journal of Cheminformatics}}
  \bibinfo{volume}{12}, \bibinfo{number}{1} (\bibinfo{year}{2020}),
  \bibinfo{pages}{1--18}.
\newblock


\bibitem[\protect\citeauthoryear{Wu, Pan, Chen, Long, Zhang, and Philip}{Wu
  et~al\mbox{.}}{2020}]%
        {wu2020comprehensive}
\bibfield{author}{\bibinfo{person}{Zonghan Wu}, \bibinfo{person}{Shirui Pan},
  \bibinfo{person}{Fengwen Chen}, \bibinfo{person}{Guodong Long},
  \bibinfo{person}{Chengqi Zhang}, {and} \bibinfo{person}{S~Yu Philip}.}
  \bibinfo{year}{2020}\natexlab{}.
\newblock \showarticletitle{A comprehensive survey on graph neural networks}.
\newblock \bibinfo{journal}{\emph{IEEE transactions on neural networks and
  learning systems}} (\bibinfo{year}{2020}).
\newblock


\bibitem[\protect\citeauthoryear{Wu, Ramsundar, Feinberg, Gomes, Geniesse,
  Pappu, Leswing, and Pande}{Wu et~al\mbox{.}}{2018}]%
        {wu2018moleculenet}
\bibfield{author}{\bibinfo{person}{Zhenqin Wu}, \bibinfo{person}{Bharath
  Ramsundar}, \bibinfo{person}{Evan~N Feinberg}, \bibinfo{person}{Joseph
  Gomes}, \bibinfo{person}{Caleb Geniesse}, \bibinfo{person}{Aneesh~S Pappu},
  \bibinfo{person}{Karl Leswing}, {and} \bibinfo{person}{Vijay Pande}.}
  \bibinfo{year}{2018}\natexlab{}.
\newblock \showarticletitle{MoleculeNet: a benchmark for molecular machine
  learning}.
\newblock \bibinfo{journal}{\emph{Chemical science}} \bibinfo{volume}{9},
  \bibinfo{number}{2} (\bibinfo{year}{2018}), \bibinfo{pages}{513--530}.
\newblock


\bibitem[\protect\citeauthoryear{Xinyi and Chen}{Xinyi and Chen}{2018}]%
        {xinyi2018capsule}
\bibfield{author}{\bibinfo{person}{Zhang Xinyi} {and} \bibinfo{person}{Lihui
  Chen}.} \bibinfo{year}{2018}\natexlab{}.
\newblock \showarticletitle{Capsule graph neural network}. In
  \bibinfo{booktitle}{\emph{International conference on learning
  representations}}.
\newblock


\bibitem[\protect\citeauthoryear{Xu, Hu, Leskovec, and Jegelka}{Xu
  et~al\mbox{.}}{2018}]%
        {xu2018powerful}
\bibfield{author}{\bibinfo{person}{Keyulu Xu}, \bibinfo{person}{Weihua Hu},
  \bibinfo{person}{Jure Leskovec}, {and} \bibinfo{person}{Stefanie Jegelka}.}
  \bibinfo{year}{2018}\natexlab{}.
\newblock \showarticletitle{How powerful are graph neural networks?}
\newblock \bibinfo{journal}{\emph{arXiv preprint arXiv:1810.00826}}
  (\bibinfo{year}{2018}).
\newblock


\bibitem[\protect\citeauthoryear{Yang, Swanson, Jin, Coley, Eiden, Gao,
  Guzman-Perez, Hopper, Kelley, Mathea, et~al\mbox{.}}{Yang
  et~al\mbox{.}}{2019}]%
        {yang2019analyzing}
\bibfield{author}{\bibinfo{person}{Kevin Yang}, \bibinfo{person}{Kyle Swanson},
  \bibinfo{person}{Wengong Jin}, \bibinfo{person}{Connor Coley},
  \bibinfo{person}{Philipp Eiden}, \bibinfo{person}{Hua Gao},
  \bibinfo{person}{Angel Guzman-Perez}, \bibinfo{person}{Timothy Hopper},
  \bibinfo{person}{Brian Kelley}, \bibinfo{person}{Miriam Mathea},
  {et~al\mbox{.}}} \bibinfo{year}{2019}\natexlab{}.
\newblock \showarticletitle{Analyzing learned molecular representations for
  property prediction}.
\newblock \bibinfo{journal}{\emph{Journal of chemical information and
  modeling}} \bibinfo{volume}{59}, \bibinfo{number}{8} (\bibinfo{year}{2019}),
  \bibinfo{pages}{3370--3388}.
\newblock


\bibitem[\protect\citeauthoryear{Ying, You, Morris, Ren, Hamilton, and
  Leskovec}{Ying et~al\mbox{.}}{2018}]%
        {ying2018hierarchical}
\bibfield{author}{\bibinfo{person}{Rex Ying}, \bibinfo{person}{Jiaxuan You},
  \bibinfo{person}{Christopher Morris}, \bibinfo{person}{Xiang Ren},
  \bibinfo{person}{William~L Hamilton}, {and} \bibinfo{person}{Jure Leskovec}.}
  \bibinfo{year}{2018}\natexlab{}.
\newblock \showarticletitle{Hierarchical graph representation learning with
  differentiable pooling}.
\newblock \bibinfo{journal}{\emph{arXiv preprint arXiv:1806.08804}}
  (\bibinfo{year}{2018}).
\newblock


\bibitem[\protect\citeauthoryear{Yuan, Wang, Coghill, and Pang}{Yuan
  et~al\mbox{.}}{2021c}]%
        {yuan2021novel}
\bibfield{author}{\bibinfo{person}{Yingfang Yuan}, \bibinfo{person}{Wenjun
  Wang}, \bibinfo{person}{George~M Coghill}, {and} \bibinfo{person}{Wei Pang}.}
  \bibinfo{year}{2021}\natexlab{c}.
\newblock \showarticletitle{A novel genetic algorithm with hierarchical
  evaluation strategy for hyperparameter optimisation of graph neural
  networks}.
\newblock \bibinfo{journal}{\emph{arXiv preprint arXiv:2101.09300}}
  (\bibinfo{year}{2021}).
\newblock


\bibitem[\protect\citeauthoryear{Yuan, Wang, and Pang}{Yuan
  et~al\mbox{.}}{2021a}]%
        {yuan2021genetic}
\bibfield{author}{\bibinfo{person}{Yingfang Yuan}, \bibinfo{person}{Wenjun
  Wang}, {and} \bibinfo{person}{Wei Pang}.} \bibinfo{year}{2021}\natexlab{a}.
\newblock \showarticletitle{A Genetic Algorithm with Tree-structured Mutation
  for Hyperparameter Optimisation of Graph Neural Networks}.
\newblock \bibinfo{journal}{\emph{arXiv preprint arXiv:2102.11995}}
  (\bibinfo{year}{2021}).
\newblock


\bibitem[\protect\citeauthoryear{Yuan, Wang, and Pang}{Yuan
  et~al\mbox{.}}{2021b}]%
        {yuan2021systematic}
\bibfield{author}{\bibinfo{person}{Yingfang Yuan}, \bibinfo{person}{Wenjun
  Wang}, {and} \bibinfo{person}{Wei Pang}.} \bibinfo{year}{2021}\natexlab{b}.
\newblock \showarticletitle{A Systematic Comparison Study on Hyperparameter
  Optimisation of Graph Neural Networks for Molecular Property Prediction}.
\newblock \bibinfo{journal}{\emph{arXiv preprint arXiv:2102.04283}}
  (\bibinfo{year}{2021}).
\newblock


\end{thebibliography}

% \appendix









\end{document}